\documentclass[conference]{IEEEtran}
\IEEEoverridecommandlockouts
\usepackage{cite}
\usepackage{amsmath,amssymb,amsfonts}
\usepackage{algorithmic}
\usepackage{graphicx}
\usepackage{textcomp}
\usepackage{xcolor}
\usepackage{mathtools}
\usepackage{commath}
\usepackage[hyphens]{url}
\usepackage{subcaption}
\usepackage{balance}

\begin{document}
% \bstctlcite{IEEEexample:BSTcontrol}

% \title{SkillRec: Occupational Skill Recommendation using Job Title Embeddings}
\title{SkillRec: A Data-Driven Approach to Job Skill Recommendation for Career Insights}
%\title{Data-Driven Artificial Intelligence Recommendation Engine for Career Insights}

\author{
\IEEEauthorblockN{Xiang Qian Ong}
\IEEEauthorblockA{\textit{Information Systems Technology and Design Pillar} \\
\textit{Singapore University of Technology and Design}\\
Singapore \\
xiangqian\_ong@alumni.sutd.edu.sg}
\and
\IEEEauthorblockN{Kwan Hui Lim}
\IEEEauthorblockA{\textit{Information Systems Technology and Design Pillar} \\
\textit{Singapore University of Technology and Design}\\
Singapore \\
kwanhui\_lim@sutd.edu.sg}
}

% for IEEE copyright statement
% 978-1-7281-6251-5/20/$31.00 ©2020 IEEE
%\IEEEoverridecommandlockouts
% \IEEEpubid{\makebox[\columnwidth]{978-1-6654-3902-2/21/\$31.00~\copyright~2021 IEEE \hfill} \hspace{\columnsep}\makebox[\columnwidth]{ }}

% make the title area
\maketitle

% for IEEE copyright statement
\IEEEpubidadjcol

\begin{abstract}
Understanding the skill sets and knowledge required for any career is of utmost importance, but it is increasingly challenging in today’s dynamic world with rapid changes in terms of the tools and techniques used. Thus, it is especially important to be able to accurately identify the required skill sets for any job for better career insights and development. In this paper, we propose and develop the Skill Recommendation (SkillRec) system for recommending the relevant job skills required for a given job based on the job title. SkillRec collects and identify the skill set required for a job based on the job descriptions published by companies hiring for these roles. In addition to the data collection and pre-processing capabilities, SkillRec also utilises word/sentence embedding techniques for job title representation, alongside a feed-forward neural network for job skill recommendation based on the job title representation. Based on our preliminary experiments on a dataset of 6,000 job titles and descriptions, SkillRec shows a promising performance in terms of accuracy and F1-score.
\end{abstract}


%%%%%%%%%%%%%%%%%%%%%%%%%%%%%%%%%%%%%%%%%%%%%%%%%%%%
%%%%%%%% Introduction
%%%%%%%%%%%%%%%%%%%%%%%%%%%%%%%%%%%%%%%%%%%%%%%%%%%%
\section{Introduction}

Technological advancements and industrial changes are constantly evolving at a rapid pace~\cite{khaouja2021survey}. With this rapid change, new job scopes are constantly being generated at companies along with a dynamic list of skill sets to meet these ever-growing changes. As such, individuals that are looking to change their career or seeking a new job often face the problem of skill mismatch due to those constant changes. This problem is further highlighted by the recent COVID-19 pandemic which has caused drastic changes in many industries, such as aviation and tourism. During this period, there was a spike in the number of retrenched workers seeking to find employment in other fields~\cite{world2020future} and potentially facing skill mismatch.

On top of that, jobs are tightly coupled with skill sets, which are essential for fulfilling the job requirements. Individuals who are transiting into a new career and industry would need to conduct extensive research or seek advice on the latest job requirements and the relevant skills. This process is commonly performed to improve the chances of applicants securing a new job, especially in a new industry. 
This is necessary due to the huge barrier of entry for individuals transiting over to new jobs or new industries, given their existing skill sets. Different industry fields and even roles within the same industry require different sets of skills to carry out tasks associated with the job. Likewise, hiring managers also look out for those skill sets when assessing suitable candidates for an open role. As such, identifying the skill sets required for a given job is especially important for potential job seekers, even for those staying in the same industry.

\subsection{Related Work}

In terms of academic research, this problem and research area have also garnered keen interest in recent years due to these various developments~\cite{gugnani2020implicit,khaouja2021survey,liu2022Title2vec,ao2023skill,barducci2022end,liu2020ipod}. For example, there are various streams of research that investigate diverse but related topics such as: (i) recommending job candidates to companies based on their suitability~\cite{zhu2018person,shen2018joint,qin2018enhancing}; (ii) recommending jobs to candidates based on the candidates skill sets and career history~\cite{malinowski2006matching,zhang2014research}; (iii) modelling the career trajectories of individuals~\cite{liu2016fortune,mimno2008modeling}; (iv) predicting the future career transitions and movements of workers~\cite{james2018prediction,yang2018one}.

Among these works, the most closely related works to our research are those on job skill recommendations~\cite{dave2018combined,maurya2017bayesian,sun2021cost} and those using various embedding techniques~\cite{liu2022Title2vec,giabelli2021skills2graph,giabelli2021skills2job,jaramillo2020word} for occupational analysis and mining. For example, Dave et al. proposed a representation learning model using job transitions, job-skills and skill co-occurence networks for recommending both jobs and skills~\cite{dave2018combined}. Others have used reinforcement learning in the form of a Deep-Q Network for skill recommendation and better interpretation~\cite{sun2021cost}. Liu et al. developed a bi-directional language model for contextualised job tile embeddings that was demonstrated on a downstream occupational named entity recognition task~\cite{liu2022Title2vec}. Skills2Graph~\cite{giabelli2021skills2job} used FastText embeddings to better maintain semantic similarity between taxonomic components of job postings. Also closely related to our research are various works that have developed systems and implemented approaches for related occupational analysis and mining tasks~\cite{ashok2018jobsense,giabelli2021skills2graph,qin2019duerquiz}.

\subsection{Motivation and Problem Statement}

The fundamental problem is that identifying the relevant skill sets required for a particular job title is both time-consuming and a complex process. However, this process is essential to job seekers looking to enter new jobs and yet very challenging as they are new to both the field and industry. The problem can be broken down into two sub-problems: the lack of knowledge to learn the relevant skills and the lack of proficiency in the relevant skills, which will be elaborated next.

\subsubsection{Lack of knowledge to learn the relevant skills}

To learn the relevant skills required for the particular job, there is a need to understand the current job workflow, the tools that are used on the job and knowledge in that field of interest. The root of the problem is the lack of domain expertise in the new industry to truly understand the skill sets required for the job~\cite{restrepo2015skill}. 

There are various different ways in which the problem can be mitigated. The first is to seek advice from experts who have worked in the field for a long time. They understand the job requirements and the necessary skill sets needed, from their extensive years of experience. Alternatively, governments’ skills recommendations serve as a good indicator of possible general relevant skill sets that are required in the industry for the near future~\cite{felstead2007skills,felstead2013skills}. Lastly, individuals could perform online research on the industry, given that there is an abundance of articles and forums that share insights and recommendations.

This step is crucial as it serves as the basis for the next step. Only after knowing the skills that need to be learned, then the individual can embark on the next step.

\subsubsection{Lack of proficiency in the relevant skills}

After identifying the required skill sets to transit to a new industry or role, it is crucial for individuals to have access to resources to learn those skills. Proficiency in these relevant skills is an essential requirement that leads to successful employment and subsequent progression. The resources could come in many different forms, both physical and digital, which we discuss next.

A classroom setting is a common structure whereby individuals will attend classes and undergo a structured curriculum to gain knowledge and experience from the instructors. At the end of such courses, there are formal certificates that act as proof of completion of the course. Classroom formats could also be virtual, ranging from online courses such as Udemy and Coursera. Similarly, these classes provide both the content and practical lessons, with the added bonus of online certificates upon completion. Openly available resources are unstructured curricula that could comprise different resources from both online and offline. This gives the individual the flexibility of time and learning opportunity but typically do not result in a formal certification upon completion.

\subsection{Problem Definition}

In this paper, our main focus is on the first sub-problem, which is the lack of knowledge to learn the relevant skills required for a job role. This problem is significantly more crucial than the second sub-problem, as the former's outcome will determine the skill sets that an individual is required to learn. Failing at this step will render useless any future efforts at both a industry transition or career progression. In addition, there are multiple open-source and affordable resources online to address the problem of a lack of proficiency, thus this second sub-problem is thus not as crucial as compared to the first sub-problem that we focus on. More formally, we investigate the problem of identifying and recommending the required job skill sets required for a role given the job title.

\subsection{Main Contributions}
% 1.4 Proposed Solution

The current ways to address the problems are mainly human-driven insights, such as seeking the advice of current practitioners in the new industry or through self-reading of openly available resources. However, we could tap into data-driven insights that perform the analysis and recommend the right skill sets to pick up for the industry or role we are interested in. This could reduce the time taken or the cost incurred in the process.

In this paper, we make the following contributions:
\begin{enumerate}
    \item We investigate the problem of job skill recommendation, which is to recommend a list of required skill sets given role in the form of a job title.
    \item To address the above problem, we propose and develop the Skill Recommendation (SkillRec) system. SkillRec comprises the capability to collect openly available job listing information, process these data to extract job title and required skill sets, and subsequently recommend skill sets given a job title.
    \item As part of SkillRec, we also develop two approaches to skill recommendation, utilising word/sentence embedding techniques (using BERT and FastText) for job title representation, before inputing this job title representation to a simple but effective feed-forward neural network for job skill recommendation.
    \item Using a dataset of 6,000 job listings, we perform preliminary experiments to evaluate the performance of our proposed SkillRec system. Our preliminary results show promising performance of SkillRec in terms of accuracy and F1-score.
\end{enumerate}

\begin{figure*}[th]
    \centering
    \includegraphics[width=0.82\linewidth, trim=0mm 0mm 0mm 0mm]{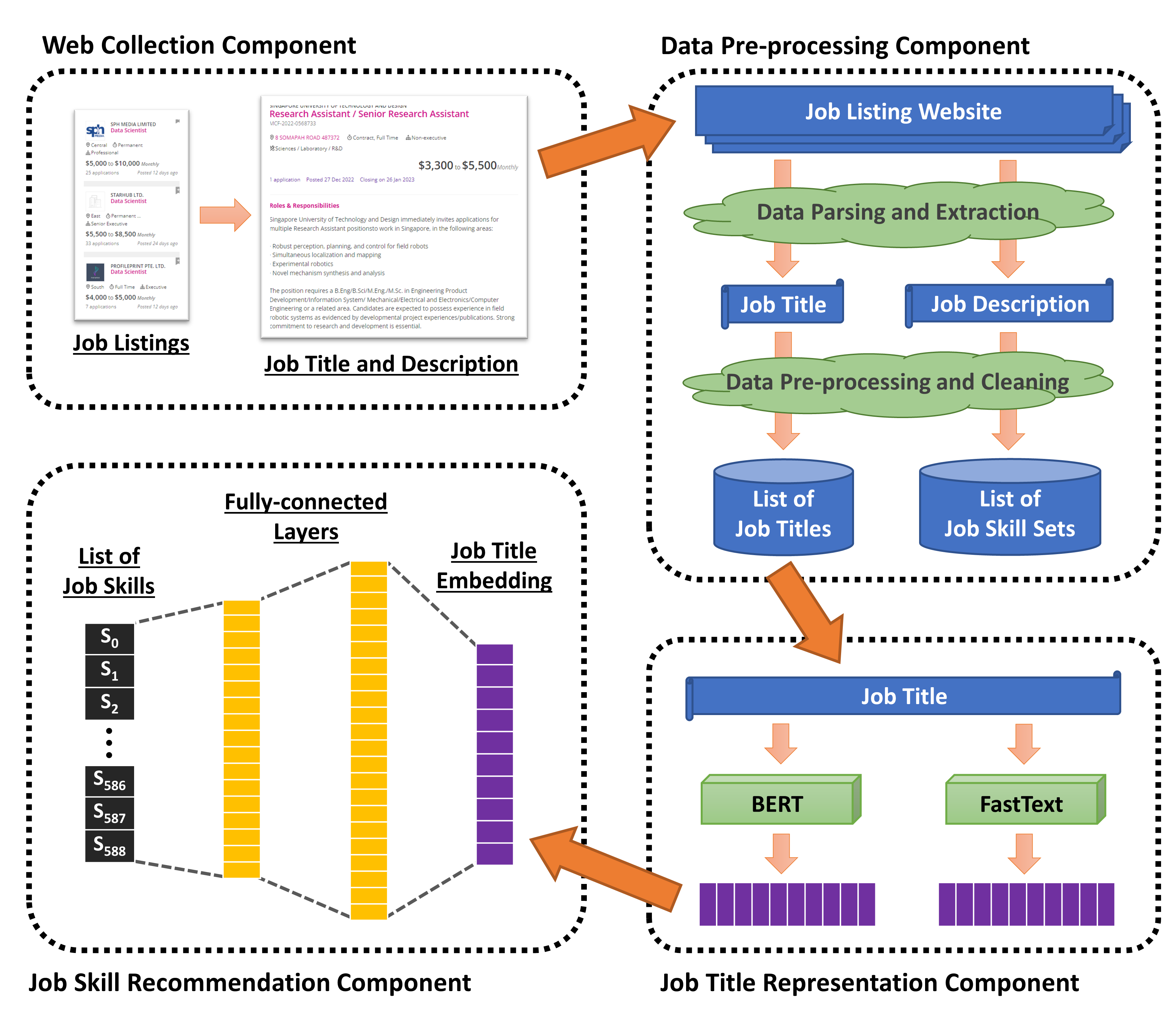}
    \caption{System Architecture of Our Proposed Skill Recommendation (SkillRec) System}
    \label{figFramework}
\end{figure*}

%Given our technology today, it is more than possible to churn out data models from user-generated data. An example would be using the skill sets required for a given job title. The user-generated data can come in the form of job descriptions published on job posting websites that are ready for consumption, such as Glassdoor and Indeed.
%Having data-driven insights will drastically reduce the cost and time required for individuals to understand the skillsets they need to learn.
%The final solution will come in the form of an Artificial Intelligence (AI) recommendation engine that produces career insights, based on public data. This solution will be built on web crawling technologies, a data processing pipeline, and a data model that will act as the recommendation engine.

%%%%%%%%%%%%%%%%%%%%%%%%%%%%%%%%%%%%%%%%%%%%%%%%%%%%
%%%%%%%% Proposed System
%%%%%%%%%%%%%%%%%%%%%%%%%%%%%%%%%%%%%%%%%%%%%%%%%%%%
\section{Proposed Skill Recommendation System}
\label{proposedSystem}

Our proposed Skill Recommendation (SkillRec) system comprises four main components, including a web collection component, data pre-processing component, job title representation component and job skill recommendation component. Figure~\ref{figFramework} shows an overview of our proposed system. Next, we provide a brief summary of each component, followed by a more detail description in the later sections. 

\begin{enumerate}
\item Web Collection Component. This component retrieves job listings from various web sources, such as online job posting platforms.
\item Data Pre-processing Component. This component performs the necessary data pre-processing and cleaning to extract relevant information from the job descriptions, such as the job title, job description, required skills, company name, etc.
\item Job Title Representation Component. This component utilises word/sentence embedding techniques, such as BERT and FastText, to model job titles in term of its vector representations.
\item Job Skill Recommendation Component. The job title representation from the previous step is then fed into this component, which comprises a feed-forward neural network that outputs a set of recommended skills given this job title.
\end{enumerate}

\subsection{Web Collection Component}

The web collection component retrieves various types of web-based information such as job descriptions from popular online job posting platforms, as well as skill sets from massive open online courses platforms. 

The job description data was retrieved over a period of six months at a weekly interval to ensure that the dataset is kept up-to-date. A total of 6,000 job descriptions were retrieved from three popular online job posting platform.

Massive open online courses platforms provide a good source of information on popular educational courses and the skill sets they aim to impart. Using two popular massive open online courses platforms, we collected and identified 589 unique skill sets that are relevant for this work.

\subsection{Data Pre-processing Component}

The data pre-processing component works on the job title and job description data collected by the web collection component. This component first parses the collected webpage to extract the relevant information, i.e., the job title and job description. Thereafter, the standard text pre-processing steps, such as conversion to lowercase, removal of non-textual symbols, tokenisation, are performed on these two fields. The skill sets for a job listing is derived from its corresponding job description based on an exact match of a mentioned skill with our earlier list of 589 unique skill sets.

\subsection{Job Title Representation Component}
\label{jobEmbedding}

As our main task is to determine the required skill sets for a given job title, we proceed to model the job titles using various embedding techniques. In this study, we experiment with two popular word/sentence representation techniques:

\begin{itemize}
    \item BERT Embedding~\cite{devlin2019bert}. The Bidirectional Encoder Representation of Transformer (BERT) is a transformer-based language model comprising multiple layers of bidirectional Transformer encoders, along with self-attention mechanism. BERT is then pre-trained with the Masked Language Model task of predicting hidden tokens using other non-hidden tokens, and/or Next Sentence Prediction task of predicting the next sentence given an input sentence.
    \item FastText Embedding~\cite{bojanowski2017enriching,joulin2017bag}. FastText builds upon the skip-gram approach that is popularly used by the Word2Vec model~\cite{mikolov2013efficient,mikolov2013distributed}. Using FastText, each word is represented by a bag-of-character n-grams, and in turn each character n-grams is represented by a vector. The vector representation of each word is then derived from the sum of these character n-grams vector representations.
\end{itemize}

BERT was chosen due to its strong performance in various language modelling tasks, such as text classification, question answering, among others. In addition, BERT has been frequently used in a variety of other recommendation and prediction tasks, such as location prediction~\cite{li2022transformer,simanjuntak2022we} and tourism recommendation~\cite{arreola2021embeddings,ho2022poibert}.

After extracting the job title in the previous step, either BERT or FastText is then used to obtain the vector representation of the job title. This job title representation is subsequently used as input to our neural network for job skill recommendation, which we describe next.

\subsection{Job Skill Recommendation Component}
\label{skillRecom}

Using the vector representation of the job title from the previous step, this job title representation is then fed into a feed-forward neural network for job skill recommendation based on the job title. We utilise a feed-forward neural network with two fully-connected hidden layer, with the first hidden layer comprising 1280 nodes, the second hidden layer comprising 640 nodes and a final output layer of 589 nodes corresponding to the 589 unique skill sets. We employ Sigmoid as our activation function, AdamW as our optimizer and a dropout of 0.5 to reduce overfitting.

%%%%%%%%%%%%%%%%%%%%%%%%%%%%%%%%%%%%%%%%%%%%%%%%%%%%
%%%%%%%% Experimental Results
%%%%%%%%%%%%%%%%%%%%%%%%%%%%%%%%%%%%%%%%%%%%%%%%%%%%
\section{Experiments and Results}

Using our collected dataset, we use 80\% for training our model as described in Section~\ref{proposedSystem} and use the remaining 20\% for evaluation. For evaluation purposes, we use the standard metrics of accuracy and F1-score, based on the recommended job skill against the ground truth job skill in the original job description.

We experimented with the following algorithms and baselines:
\begin{itemize}
    \item Bag-of-Words. A classical bag-of-words model that is traditionally used for classification tasks involving text-based data. In our application, we utilise a frequency-based approach to model the frequency of which a particular skill is mentioned in relation to a specific job. This model is used as a simple baseline to compare our two proposed approaches.
    \item FastText+NN. As described earlier in our SkillRec system, this model uses the FastText embedding for job title representation (Section~\ref{jobEmbedding}), followed by a neural network (NN) for job skill recommendation (Section~\ref{skillRecom}).
    \item BERT+NN. Similar to the earlier FastText+NN model, this model uses the BERT embedding for job title representation (Section~\ref{jobEmbedding}), followed by the same neural network (NN) for job skill recommendation (Section~\ref{skillRecom}).
\end{itemize}
    
\begin{table}[!thbp]
\renewcommand*{\arraystretch}{1.25}
\centering
\caption{Experimental Results}
\label{expResults}
%\resizebox{0.8\textwidth}{!}{
\begin{tabular}{lll} 
\hline 
Algorithm & Accuracy & F1-score \\
\hline  
Bag-of-Words & 0.4768 & 0.2887 \\
FastText+NN & 0.9728 & 0.4931 \\
BERT+NN & 0.9870 & 0.4973 \\
\hline
\end{tabular}
%}
\end{table}

Table~\ref{expResults} shows the average results of our experiments for the three models in terms of accuracy and F1-score. The results show that our proposed SkillRec system using BERT+NN is the best performing model, with FastText+NN being a extremely close second. BERT has been shown to perform well for a range of natural language processing tasks, from text classification to question answering. Similarly for this task of occupational skill recommendation, BERT as well as FastText perform well with the add-on of a straightforward neural network. Unsurprisingly, Bag-of-Words performs the worst by a large margin due to the simple job title representation approach that is unable to capture similar job titles obtained.

\begin{table}[!thbp]
\renewcommand*{\arraystretch}{1.25}
\centering
\caption{Example Job Title and Relevant Skills}
\label{exampleRecom}
%\resizebox{0.8\textwidth}{!}{
\begin{tabular}{ll} 
\hline 
Job Title & Related Jobs and Skills \\
\hline  
developer	& java, python, fullstack, android, software \\
software	& stack, java, python, fullstack, android \\
chef	& pastry, kitchen, hair, chemist, accountant \\
senior director	& director, partner, pacific, planner, strategy \\
software engineer	& software, engineer, python, java, embedded \\
\hline
\end{tabular}
%}
\end{table}

In addition to the experiments measuring accuracy and F1-score of the job skill recommendations, we also performed a qualitative check on the job title representations learnt and the similar job titles and skill sets based on these representations. Table~\ref{exampleRecom} shows a sample of some job titles and the related job titles and skill sets.
This simple analysis shows the effectiveness of our approach as occupations such as developer and software engineer results in relevant skills like Java, Python, fullstack, embedded, etc. Similarly, non-technical roles such as senior director also resulted in similar roles like partner and planner and relevant skills like strategy. For certain roles like chef, the results are more mixed with relevant terms like pastry and kitchen being mentioned but also non-relevant ones like chemist and accountant. This issue highlights a limitation due to our preliminary experiments and limited dataset, which contains more technical roles than non-technical ones. 

%%%%%%%%%%%%%%%%%%%%%%%%%%%%%%%%%%%%%%%%%%%%%%%%%%%%
%%%%%%%% Conclusion and Future Work
%%%%%%%%%%%%%%%%%%%%%%%%%%%%%%%%%%%%%%%%%%%%%%%%%%%%
\section{Conclusion and Future Work}

In this paper, we proposed the SkillRec system to address the problem of job skill recommendation given a job title. SkillRec comprises four main components for web collection, data pre-processing, job title representation and job skill recommendation. In particular, our preliminary experiments show that an approach that combines job title representation using BERT, along with job skill recommendation using a straightforward feed-forward neural network works effectively in terms of accuracy and F1-score.

Future enhancements to SkillRec can involve a more dynamic recommendation of job skills to account for new skill sets that may emerge in future or existing skill sets that evolve over time. Similarly, we can also include more context in addition to the job titles, such as the industry, country, job applicant details, etc, for making more contextualised and personalised job skill recommendations.

\section{Acknowledgment}
This research is funded in part by the Singapore University of Technology and Design under grant RS-MEFAI-00005-R0201.

\balance
\bibliographystyle{IEEEtran}
% argument is your BibTeX string definitions and bibliography database(s)
\bibliography{careerRecom}

\end{document}